%% file: main.tex
\documentclass[10pt]{article} 
\usepackage[preprint]{tmlr}

\input{math_commands.tex}

\usepackage{hyperref}
\usepackage{url}

\usepackage{caption}
\usepackage{amsmath}

\usepackage{tikz}
\usepackage{pgfplots}
\usepackage{pgfplotstable}
\usetikzlibrary{shapes.geometric, arrows}
\usetikzlibrary{pgfplots.statistics, pgfplots.colorbrewer}  
\usepgfplotslibrary{groupplots}
\usetikzlibrary{shapes.multipart}
\usetikzlibrary{patterns}
\usepackage{placeins}
\usepackage{svg}

\usepackage{subcaption}
\usepackage{siunitx}
\usepackage{acro}

\usepackage[capitalise]{cleveref}
\usepackage{adjustbox}
\usepackage{booktabs}

\definecolor{b8}{RGB}{0,100,100} 
\definecolor{b16}{RGB}{30, 50, 100}
\definecolor{b32}{RGB}{100, 100, 100}
\definecolor{b64}{RGB}{100, 50, 30}
\definecolor{b128}{RGB}{200, 50, 30}

\newcommand\nn{M}

\definecolor{baseline1}{RGB}{180, 40, 200}
\definecolor{baseline2}{RGB}{0, 0, 140}
\definecolor{fastest}{RGB}{200, 20, 20}
\definecolor{ours}{RGB}{100, 200, 100}

\DeclareAcronym{VFS}{short=VFS, long=Voltage Frequency Scaling}
\DeclareAcronym{SOC}{short=SOC, long=System On Chip}
\DeclareAcronym{NN}{short=NN, long=Neural Network}
\DeclareAcronym{DVFS}{short=DVFS, long=Dynamic Voltage and Frequency Scaling}

\usepackage{algorithm}
\usepackage[noend]{algpseudocode}

\usepackage{comment}

\title{Accelerated Training on Low-Power Edge Devices}

\author{\name Mohamed Aboelenien Ahmed \email mohamed.ahmed3@kit.edu \\
      \addr Karlsruhe Institute of Technology
      \AND
      \name Kilian Pfeiffer \email kilian.pfeiffer@kit.edu \\
      \addr Karlsruhe Institute of Technology
      \AND
      \name Heba Khdr \email heba.khdr@kit.edu \\
      \addr Karlsruhe Institute of Technology \\
      \AND
      \name Osama Abboud \email osama.abboud@huawei.com \\
      \addr Huawei Research Center Munich \\
      \AND
      \name Ramin Khalili \email ramin.khalili@huawei.com \\
      \addr Huawei Research Center Munich \\
      \AND
      \name Jörg Henkel \email henkel@kit.edu \\
      \addr Karlsruhe Institute of Technology }

\def\month{MM}  
\def\year{YYYY} 
\def\openreview{\url{https://openreview.net/forum?id=XXXX}} 

\begin{document}

\maketitle

\begin{abstract}
Training on edge devices poses several challenges as these devices are generally resource-constrained, especially in terms of power.  
State-of-the-art techniques at the device level reduce the GPU frequency to enforce power constraints, leading to a significant increase in training time. To accelerate training, we propose to jointly adjust the system and application parameters (in our case, the GPU frequency and the batch size of the training task) while adhering to the power constraints on devices. 
We introduce a novel cross-layer methodology that combines predictions of batch size efficiency and device profiling to achieve the desired optimization. 
Our evaluation on real hardware shows that our method outperforms the current baselines that depend on state of the art techniques, reducing the training time by $2.4\times$ with results very close to optimal. Our measurements also indicate a substantial reduction in the overall energy used for the training process. These gains are achieved without reduction in the performance of the trained model. 
\end{abstract}

\input{introduction}
\input{background}

\input{problem_statement}

\input{methodology}

\input{results}

\input{conclusion}

\bibliography{bibliography}
\bibliographystyle{tmlr}

\input{appendix}

\end{document}

%% file: math_commands.tex
\usepackage{amsmath,amsfonts,bm}

\def\eqref#1{equation~\ref{#1}}

\def\1{\bm{1}}

\DeclareMathAlphabet{\mathsfit}{\encodingdefault}{\sfdefault}{m}{sl}
\SetMathAlphabet{\mathsfit}{bold}{\encodingdefault}{\sfdefault}{bx}{n}



%% file: introduction.tex
\section{Introduction}

Recent trends are moving toward the deployment of deep learning models on edge devices due to the need for real-time decision-making, reduced communication cost, and privacy compared to offloading computation to cloud-based servers \cite{object_detection_on_mobile, dl_on_mobile, dl_on_edge_survey, object_detection_edge}. Low-latency inference is essential in applications such as autonomous vehicles, robotics, surveillance, and smart agriculture. To meet these requirements, edge devices are equipped with compact GPUs optimized for computer vision and deep learning inference tasks \cite{embedded_gpus, pedestrian_gpus, agriculture_gpus, object_tracking, lstm_mobile_gpus, object_detection_edge}. Such devices are often power constrained, running on batteries with limited power outputs.

Many applications on edge devices involve sensitive or private information that cannot be shared due to regulatory compliance or ethical concerns, making it challenging to maintain accurate models without access to this data. 
Training on the edge offers adaptability to the data available on these devices while reducing the risk of exposure during transmission. Transfer learning is thereby applied for on-device training, where a model is pre-trained with a large (public) dataset and fine-tuned on-device on the target domain \cite{tinyTL, MobileTL, under_256k_memory, on_device_gradient_filtering}. While this offers less training time compared to training from scratch, transfer learning is still a computationally demanding task utilizing higher power compared to inference.

To accelerate training, usually large batch sizes for mini-batch gradient descent are used. However, increasing the batch size increases the power consumption which is limited at edge devices. To enforce the power constraint, the circuit-level control will select the frequency level that satisfies this constraint considering the worst-case computation scenario. Intuitively, this will decelerate training: in the state of the art, \emph{the system-level and application-level (training) parameters are set independently from each other}, and \emph{this is sub-optimal} as will be demonstrated in the following example. 

\begin{figure*}[tb!]
    \centering
    \includegraphics[]{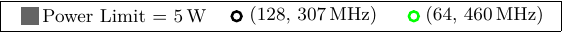}
    \begin{subfigure}[b]{0.49\textwidth}
    \centering
        \includegraphics[]{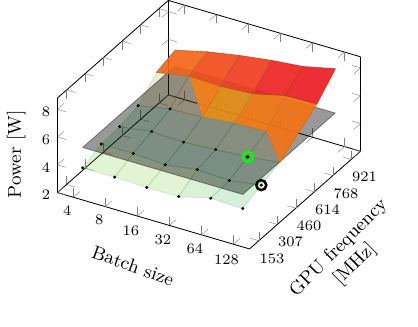}
        \caption{Power}
        \label{fig:3d_power_freq_constraint}
    \end{subfigure}
    \hfill
    \begin{subfigure}[b]{0.49\textwidth}
        \centering
        \includegraphics[]{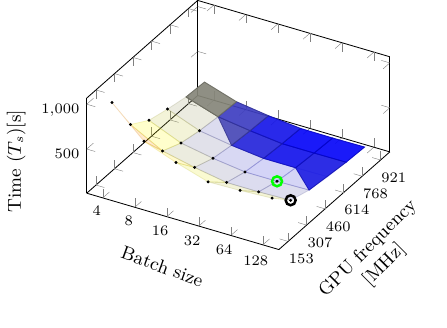}
        \caption{Time}
        \label{fig:3d_time_freq_constraint}    
    \end{subfigure}
    \caption{Peak power and time for training a set of samples across different batch size and GPU frequency combinations. The gray plane represents the power limit in the left figure, and the black dots in both figures are the feasible combinations that can be utilized under that constraint. The black circle represents the operating point with maximum feasible frequencies for the batch sizes of 128, which will be selected by the state-of-the-art techniques. The green circle represents an operating point at batch size 64, that could be selected when the frequency and the batch size are jointly selected to accelerate training under the power constraint. Selecting this operating point accelerates training by $31.9\%$ (see Time curves).}
    \label{fig:3d_profiling}
    
\end{figure*}

\Cref{fig:3d_profiling} shows the power and time required to train a ResNet18 model on Nvidia Jetson Nano with~$4096$ of CIFAR10's training samples, while using different combinations of the GPU's frequency and batch size.
Following the state of the art, and for a power constraint of~$\SI{5}{\watt}$, the system would select the GPU's frequency~$f=\SI{307}{\mega\hertz}$ and the batch size $b=128$\footnote{We select the maximum batch size which is allowed by the memory at the device.}. 
When we look at the training time, this selection is sub-optimal. In particular, we could choose a smaller batch size ($b=64$), allowing the GPU to operate at a higher frequency ($f=\SI{460}{\mega\hertz}$), and thus reducing the training time by~$31.9\%$.  
\emph{This shows that to accelerate the training on low-power edge devices, both the GPU frequency (a system parameter) and the batch size (an application parameter) should be jointly selected}. Further, a training process is more complex: a model is trained over multiple iterations until a target accuracy is reached. This adds a new dimension (accuracy) to the problem which is not included in this example. We will explore this in detail in Section~\ref{sec:problem_statement}. 

We propose a cross-layer methodology that enables joint optimization of the system and the application parameters to accelerate training under device power constraints. Our solution involves offline profiling of the \ac{NN} training on the device, measuring power and time requirements for various batch size and GPU frequency combinations. Additionally, we estimate the relationship between the batch size and the required number of training samples to reach a target accuracy through training on a proxy dataset on a server. By integrating both aspects, we select combinations that minimize the total training time while adhering to the given power constraints. Our cross-layer control of power consumption can exploit operating points from the design space which were not explored by the state of the art, significantly reducing the training time to accuracy. 
 In summary, we make the following contributions:
\begin{itemize}
    \item We demonstrate that the joint adjustment of the GPU frequency and the batch size given a power constraint is essential for accelerating on-device training.
    \item We propose a novel cross-layer methodology that aims at accelerating on-device training while adhering to the given power constraint of the device.
    \item Our evaluation on real hardware using different models (Convolution Neural Networks (CNNs) and transformers) and datasets shows a significant improvement in the training speed by over~$2.4\times$ compared to baselines that employ state-of-the-art techniques. We also observe a significant reduction in the total energy used for the training at the device, decreasing the carbon footprint of the process.  
    \item We provide a comprehensive sensitivity analysis that demonstrates the robustness of our solution towards the proxy dataset selection, validating the practicality of our approach and ensuring its effectiveness.
\end{itemize}

%% file: background.tex
\section{Background and Related Work}
\label{sec:background}
This section discusses the control parameters (i.e., GPU frequency and batch size)  used in this work and the state-of-the-art techniques.
We do not consider training on the CPU for the GPU-equipped devices, as the GPU is an order of magnitude faster (and energy efficient) while consuming nearly the same average power (see \cref{appendix:CPU_training} for some related discussion).

\subsection{\ac{VFS}} 
\label{sec:vfs}
\ac{VFS} is a system-level technique used for power management in processors~\cite{dvfs_thread_powercap, power_constrainted_gpus}. 
It adjusts the voltage and frequency at runtime, providing a trade-off between performance and energy efficiency \cite{dvfs_energy, dvfs_gpu_energy}. Reducing each of the processor's frequency~$f$ and voltage~$V$ values can result in a cubic reduction in the dynamic power, i.e., 
$
    P_{\text{dynamic}} \approx fV^{2}.
$
Within a device, the chip manufacturer provides a discrete set of frequencies to operate with, where~$f_{i}$,  $i \in \{1, 2, \ldots , \text{max}\}$. 
The voltage is a function of the operating frequency~$f$, and will be updated automatically for any given frequency. 
To enforce power constraints on processors, the industry standard technique is to have an upper bound $f$ (and $V$) for the device to use under a power limit. \ac{VFS} is also applied to the CPU's frequency, but this is beyond the scope of our work as we train on the GPUs.

\ac{VFS} has been studied in \cite{dvfs_precision_control_nn, dvfs_energy_efficiency_hw_nn} for improving performance and energy efficiency for inference only. Tang et al. \cite{gpu_dvfs_impact_study} conducted a study on the impact of GPU \ac{VFS} on performance and energy for both training and inference. However, the study mainly focused on the computational perspective without addressing accuracy and training speed till convergence. In contrast, we show how changing the frequency and batch size due to power limits, accompanied by the difference in iterations to reach accuracy, can lead to different optimal configurations in training.

\subsection{Batch Size} 
One of the state-of-the-art techniques for training \acp{NN} is the minibatch gradient descent, where gradients are computed using the samples drawn in mini-batch to approximate the overall gradient, enabling iterative updates of the model parameters. 
The latency for processing a batch is influenced by the GPU's ability to parallelize computations and its memory bandwidth. 
Increasing the batch size generally decreases the latency of processing data samples by parallelizing more computations on multiple processing units. 
Nevertheless, this increases the number of computational operations on the GPU, and hence the power consumption. 

The impact of batch size on accuracy and convergence speed is explored from multiple aspects. \cite{goyal2017accurate, krizhevsky2014one, batch_size_and_optimizers} studied the interdependencies between batch size and other hyperparameters such as learning rate, weight decay, and optimizers. 
\cite{incraese_bs_training}  proposed to increase the batch size during training rather than decreasing the learning rate to leverage the regularization effect employed by larger batch sizes.  \cite{large_batch_size_training} proposed a gradient noise scale metric to predict the most efficient batch size that maximizes the throughput in the next training step. This approach is tailored for distributed learning, where a batch of data is split across multiple devices.
These aforementioned works have not considered the power efficiency of batching on the device. 
In contrast, the works in \cite{batch_sizer_inference,batch_dvfs_inference} consider the impact of the batch size on power consumption of inference operations. In particular, a binary search approach is employed in \cite{batch_dvfs_inference} to find an appropriate batch size that accelerates inference and then the GPU frequency is adjusted accordingly to satisfy power constraints on GPU servers. \textit{Most importantly, this approach only considers inference where the statistical impact of the batch size does not exist}.
\cite{zeus} aimed at optimizing the energy efficiency of periodic training jobs (continuously re-executed for incoming data flow) on powerful GPU clusters. To achieve their optimization (which is different from the one we target in this paper), they propose to adjust the batch size, set a power limit on the server, and depend on circuit-level control to select the frequency that satisfies that power limit.
In particular, they obtain Pareto-optimal combinations of batch sizes and power limits that optimize for energy and performance through profiling the whole training job on the server, since this job will be repeated for new data flows. However, this solution can be applied on powerful GPU clusters and not on edge devices with limited resources. 

\textit{
In summary, none of the state of the art has addressed the joint selection of the batch size and GPU frequency to accelerate training at edge devices under power constraints.}

%% file: problem_statement.tex
\section{Problem Statement}
\label{sec:problem_statement}

\begin{figure*}[tb!]
    \begin{subfigure}[b]{0.35\textwidth}
        \centering
        \includegraphics[]{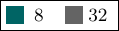}
        \includegraphics[]{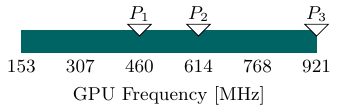}
        \includegraphics[]{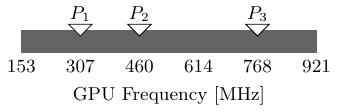}
        
    \end{subfigure}
    \hspace{6mm}
    \begin{subfigure}[b]{0.46\textwidth}
        \includegraphics[]{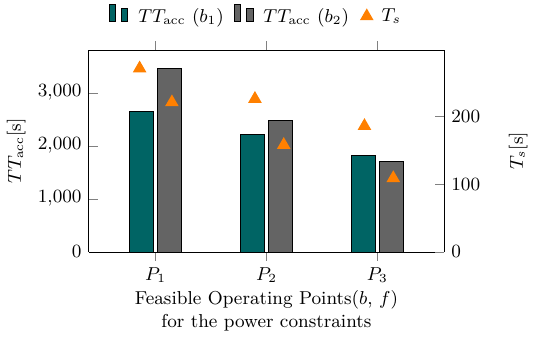}
    \end{subfigure}

    \caption{The training time on fixed number of samples~$T_s$ and the total training time~$TT_\text{acc}$ to reach an accuracy threshold of~$78\%$ using two batch sizes 8 and 32, while considering the maximum feasible GPU frequencies under three power constraints; $P_1=\SI{4.5}{\watt}$, $P_2=\SI{5}{\watt}$, and~$P_3=\SI{7}{\watt}$. We observe that for~$P_1$ and~$P_2$, selecting $b=8$ will lead to lower~$TT_\text{acc}$, while for~$P_3$ selecting $b=32$ is better. This is in contrast with our observation for~$T_s$, where selecting $b=32$ is the best option in all cases.}
    \label{fig:b8_b32_power_constraint}
\end{figure*}

We consider the following scenario: for a specific training task, an edge device requests a pre-trained \ac{NN} model~$\nn$ with its weights~$\theta$ from a server in order to fine-tune it on local data~$D$ till reaching a given accuracy threshold. 
Importantly, the edge device has a power constraint~$P_{\text{max}}$, which should not be exceeded during the training process. 
Our goal in this paper is to \textit{minimize the training (fine-tuning) time at edge devices under their given power constraints.}

 We introduce~$T_s$ as the training time required to apply training using a fixed number of samples~$s$. As shown in \cref{fig:3d_profiling}, the joint selection of $b$ and $f$ will help reduce $T_s$  under a power constraint. However, the ultimate optimization goal is to minimize the total training time required to reach a target accuracy, which we label~$TT_{\text{acc}}$.
A set of parameters, i.e., frequency and batch size~$(f,b)$, that are optimal for training a fixed number of samples ($T_s$) might not be necessarily optimal for the training to accuracy ($TT_{\text{acc}})$.

We display in~\cref{fig:b8_b32_power_constraint} the training time to reach an accuracy  of~$78\%$ (~$TT_{\text{acc}}$) for ResNet18 using two batch sizes of~$b_1=8$ and~$b_2=32$ under three different power constraints (i.e., $P_1=\SI{4.5}{\watt}$, $P_2=\SI{5}{\watt}$, and $P_3=\SI{7}{\watt}$). 
We notice that for the three power constraints, selecting~$b_1$ allows to utilize a higher frequency than~$b_2$.
For each batch size, we select the highest frequency that satisfies the power constraint, and measure~$T_s$ and~$TT_{\text{acc}}$.
We observe that using~$b_2$ (the higher batch size) always leads to a lower~$T_s$. 
However, selecting the same batch size over~$b_1$ leads to a longer~$TT_{\text{acc}}$ for~$P_1$ and~$P_2$, and shorter~$TT_{\text{acc}}$ for~$P_3$. 

This shows the complexity of the targeted problem. In particular, $TT_{\text{acc}}$ does not only depend on $T_s$, but it also depends on the number of times of processing~$s$ to reach target accuracy ($N_{{s}_\text{acc}}$). In this example, ~$N_{{s}_\text{acc}}$ for $b_2$ is equal to $15$ while ~$N_{{s}_\text{acc}}$ for $b_1$ is equal to $10$. These values and the effect of power constraint on the feasible frequency highly influence the optimal batch size to minimize~$TT_{\text{acc}}$. 
In summary, there is no clear indication on how to select the optimal operating points $(f,b)$ to achieve the target goal. 

We formulate our optimization problem as follows:
\begin{equation}
    \begin{aligned}
        & \underset{b\in\mathcal{B}, f\in\mathcal{F}}{\text{min}}
        & & TT_{\text{acc}}(b, f, \nn, D) \\
        & \text{subject to}
        & & P(b, f, \nn) \leq P_{\text{max}}
    \end{aligned}
\label{eq:optimization}
\end{equation}
where $\mathcal{B}$ is the set of feasible batch sizes, $\mathcal{F}$ is the set of available GPU's frequencies, and $P(b, f, \nn)$ is the required power to training~$M$ using~$b$ and~$f$. We rewrite ~$TT_{\text{acc}}$ as the multiplication of~$T_{s}$ and ~$N_{{s}_{\text{acc}}}$, we thus have:
\begin{equation}
    TT_{\text{acc}}(b, f, \nn, D) = T_{s}(b, f, \nn) \times  N_{{s}_{\text{acc}}}(b, \nn, D)
    \label{eq:time_to_acc2}
\end{equation}
$s$ is selected, s.t.~$b_{\text{max}} \leq s \leq |D|$, where~$b_{\text{max}}$ is the largest batch size that can fit into the memory of the devices. 
This detached formulation enables our proposed optimization method, presented in \cref{sec:methodology}.
In particular, the first factor~$T_{s}$ does not depend on the training data~$D$, nor on the accuracy threshold. The second factor~$N_{{s}_\text{acc}}$ is independent of the GPU frequency of the device.


%% file: methodology.tex
\section{Power-Aware Training}
\label{sec:methodology}

\begin{figure*}[!t]
    \centering
     \resizebox{0.9\textwidth}{0.46\textwidth}
    {

    \includegraphics{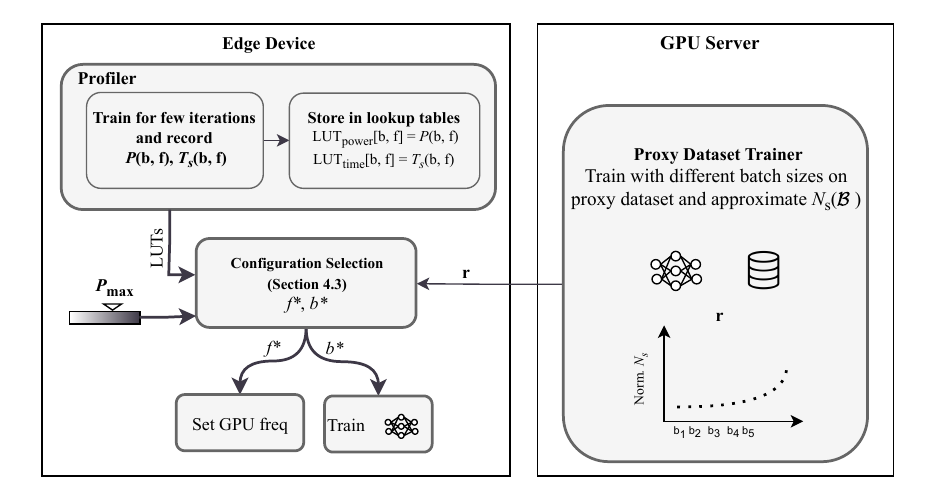}
    }
    
    \caption{Overview of our proposed cross-layer approach that accelerate training under power constraint though the joint selection of batch size $b$ and GPU frequency $f$.} 
    \label{fig:overview_figure}
\end{figure*}

We propose an optimization approach that co-selects~$b$ and~$f$ to minimize~$TT_\text{acc}$ on an edge device under~$P_\text{max}$. 
Following the problem split that we propose in \Cref{eq:time_to_acc2}, our solution consists of two main parts. The first one considers device specifics, and involves measuring time and power for training a given model~$\nn$, i.e., $T_{s}(b, f, \nn)$ and $P(b, f, \nn)$ for every $b\in\mathcal{B}$ and~$f\in\mathcal{F}$.
The second part is responsible for estimating the efficiency of batch sizes, i.e., their impact on~$N_{s_{\text{acc}}}$. This part does not depend on device characteristics and it is computationally expensive, therefore we consider it to be performed on the server that is responsible for pre-training and sending of the model. 

The workflow for our approach is as follows: The device first sends a request to the server for a pre-trained neural network model ($\nn$) with a specific architecture and input size. The server responds with the requested model and its pre-trained weights. Our proposed power-aware training then begins. The device profiles the model in terms of time and power as will be described in \cref{subsec:profiling}, while the server estimates~$N_{s_{\text{acc}}}$ for various batch sizes, as will be discussed in \Cref{subsec:proxy}, and sends these estimations to the device. Based on the profiling and estimations, the device selects the best combination of batch size and frequency to minimize~$TT_\text{acc}$. An overview is presented in \Cref{fig:overview_figure}.

\subsection{Profiling} \label{subsec:profiling}

The power and time for training on a device also depend on the \ac{NN}'s architecture~$\nn$, in addition to the frequency and batch size. Besides, changing the input shape, such as the image size in vision tasks, can lead to different time and power requirements even when using the same \ac{NN} architecture. Furthermore, the time and power are independent of the actual training data. Therefore, profiling~${T_{s}(b, f, \nn)}$ and~$P(b, f, \nn)$ given the task's input shape can be performed before full data acquisition, as long as the input dimensions of the data samples are known.

Processing a few mini-batches is adequate to obtain accurate profiling. For each batch size and frequency combination, we set the GPU frequency to~$f$ and process a few mini-batches~$m$ of size~$b$. The power sensor values are monitored to extract the peak power. Furthermore, the processing time is recorded, and then the average processing time for a mini-batch is calculated. This average processing time is then scaled to the processing time for~$s$ samples, denoted as~$T_{s}$.~$T_{s}(b, f, \nn)$ and~$P(b, f, \nn)$ are then stored in two lookup tables, denoted as~$\text{LUT}^{\nn}_{\text{Time}}$ and~$\text{LUT}^{\nn}_{\text{Power}}$.

The proposed profiling strategy considers hardware and \ac{NN} structures. Also, the time and power values are not affected by the network weight updates; thus, profiling is applicable before receiving pre-trained weights if~$\nn$ is available at the device and is needed to be perdomed only once. In a future and more practical setting, we could assume such profiling to be provided by the device manufacturer and for a pre-defined set of tasks and models.


Furthermore, if ~$P_{max}$ is known at the device, the profiling can be performed more efficiently as follows. Given that $\mathcal{B}$ and $\mathcal{F}$ are sorted, and power consumption increases with both $b$ and $f$, we begin by profiling the largest $b$ with the minimum $f$, incrementing $f$ until the maximum feasible value under ~$P_{max}$ is reached. Next, we move to the second largest $b$, starting from the highest feasible frequency found for the previous batch size, and repeat the process until the smallest batch size is profiled. This profiling is equivalent at most to 1.9\% and 6\% of the training times performed on ResNet18 and MobileNetV2 in \Cref{subsec:training_time_evaluation}.

\subsection{Estimation of~$N_{s_{\text{acc}}}$ for Batch Sizes} \label{subsec:proxy}

As discussed in \cref{sec:problem_statement}, batch sizes have different efficiency in terms of~$ N_{{s}_{\text{acc}}}(b_{i}, \nn, D)$. 
Estimating~$ N_{{s}_\text{{acc}}}(b_{i}, \nn, D)$ for every~$b_{i}\in\mathcal{B}$ depends also on the training data that is only available at the device and $\nn$'s pre-trained weights. To solve \Cref{eq:optimization}, the exact number of samples processed to reach target accuracy for each batch size can be replaced with the relative ratio between batch size~$r_{b_{i}}$ (i.e., normalized to the maximum~$N_{s_{\text{acc}}}$) as follows:
\begin{equation}
    r_{b_{i}}= \frac{N_{{s}_{\text{acc}}}(b_{i}, \nn, D)} {\underset{b \in \mathcal{B}}{\max} N_{{s}_{\text{acc}}}(b, \nn, D)}
\end{equation}
With this simplification, we adjust our focus to estimate the relation vector between batch sizes such that $\mathbf{r} = (r_{b_{1}},r_{b_{2}}, \ldots ,r_{b_\text{{max}}})$, where~$r_{b_{i}} \in (0, 1]$. However, this is still a complex task to solve given the non-linear training dynamics of deep learning; especially as the convergence speed of every batch size changes across training, making it impossible to estimate with few probes of multiple batch sizes given the training state. Obviously, training till convergence for multiple batch sizes is computationally-expensive task and cannot be conducted on the device. These all make on-device estimation of $\mathbf{r}$ inaccurate, if not infeasible.

We thus propose to estimate~$\mathbf{r}$ on a powerful GPU server. Particularly, we train~$\nn$ (with the same weights and optimizer) with multiple batch sizes until convergence (reaching the target accuracy) for a proxy dataset~$D_S$, since the server does not have an access to~$D$. This systematic exploration allows us to comprehensively assess the long-horizon impact of the batch size on the model's convergence while leveraging the computational capabilities of the server,  along with the datasets available on it and augmentation techniques.  Since~$\nn$, which would be used on the edge device, is already designed for a specific task type and pre-trained on a public dataset, a proxy dataset~$D_{S}$ should share the same task type (e.g., image object classification) and similar input shapes. Training network~$\nn$ on~$D_{S}$, despite having different objectives, allows us to estimate the relationship between batch sizes and their relative examples to accuracy on~$D$. 
Thus, we can finally have a mapping such that $\mathbf{r}_{D_{S}} \approx \mathbf{r}_{D}$.  

By estimating the batch size relation vector~$\mathbf{r}$ on the server, any edge device aiming to train~$\nn$ to utilize this vector. In \Cref{subsec:training_time_evaluation}, we provide evaluation for two different devices, namely Nvidia Jetson Nano and Nvidia Jetson TX2NX, utilizing~$\mathbf{r}$.

\begin{algorithm}[tb!]
    \caption{Batch size and GPU frequency selection}
    \label{alg:configuration_selection}
        \input{configuration_selection}

\end{algorithm}
\subsection{Batch Size and Frequency Selection}
\label{subsec:selection}
The device profiling and estimation of batch size efficiently are performed in an offline manner and at the design time. In contrast, the configuration selection is performed at runtime, as described below.
Given a power constraint~$P_{\text{max}}$, and the power measurements stored for training a specific~$\nn$ in~$\text{LUT}^{\nn}_{\text{Power}}$, we construct a set of feasible combinations~$C$ consisting of every feasible batch size with its corresponding highest (and fastest) frequency satisfying~$P_\text{max}$ as follows: 
\begin{equation}
    \begin{aligned}
    &C \gets \{(i, j_{i})) | i \in [1, \ldots, |\mathcal{B}|] \}, \\
    &j_{i} \gets \max \{j| j \in [1, \ldots, |\mathcal{F}|], \text{LUT}^{\nn}_{\text{Power}}(i, j) < P_{\text{max}}\} 
    \end{aligned}
\label{eq:feasible_combination}
\end{equation}

The processing time for the feasible combinations is then extracted from~$\text{LUT}^{\nn}_{\text{Time}}$. Following this, an approximate training time is computed by multiplying the time for every $b_{i}$ (and frequency) by the corresponding $\mathbf{r_{i}}$ element from the relation vector~$\mathbf{r}$ (i.e., estimated at design time). The configuration ($b^{*}$, $f^{*}$) that minimizes the approximate total training time is selected. Finally, we set the GPU frequency to~$f^{*}$ and then start the training using batch size~$b^{*}$.

We provide the configuration selection in \cref{alg:configuration_selection}. The selection part is $\mathcal{O}(n^{2})$ in the worse case scenario ($\mathcal{O}(|\mathcal{B}| \times |\mathcal{F}|))$. Since $\mathcal{B}$ and $\mathcal{F}$ are sorted and the same for the power and time for $|\mathcal{F}|$ for each $b$, the selection can easily be transformed to $\mathcal{O}(n\log n)$ by a binary search and is negligible to training time.

%% file: configuration_selection.tex
\begin{algorithmic}[1]
    \Require Power Limit $P_{\text{max}}$, Neural Network $\nn$, List of feasible batch sizes $\mathcal{B}$, List of GPU frequencies $\mathcal{F}$, Dataset at server $D_{S}$

    \State \textbf{Server:}
    \State $\mathbf{r} \gets \text{Proxy}(\mathcal{B}, \nn, D_{S})$ \Comment{See section (\Cref{subsec:proxy})}
    \State
    \State \textbf{Device:}
     
    \State Let $\text{LUT}^{\nn}_{\text{Time}} \in \mathbb{R}^{|\mathcal{B}| \times |\mathcal{F}|}$
    
    \State Let $\text{LUT}^{\nn}_{\text{Power}} \in \mathbb{R}^{|\mathcal{B}| \times |\mathcal{F}|}$
    \For {$i \in 0, 1, ..., |\mathcal{B}|$}
        \State $b \gets \mathcal{B}[i]$
        \For {$j \in 0, 1, ..., |\mathcal{F}|$}
                \State $f \gets \mathcal{F}[j]$ 
                \State $\text{time}, \text{power} \gets \text{Profile}(b, f, \nn)$ \Comment{On Device (\Cref{subsec:profiling})}
                \State $\text{LUT}^{\nn}_{\text{Time}}[i, j] \gets \text{time}$
                \State $\text{LUT}^{\nn}_{\text{Power}}[i, j] \gets \text{power}$
        \EndFor
    \EndFor

    \State \textit{// On-device Configuration Selection:}
    \State $TT_{\text{acc}} = \text{LUT}^{\nn}_{\text{Time}} \times \mathbf{r}$ \Comment{Element-wise multiplication} 
    
    \State $C \gets \Call{GetCombinations}{\mathcal{B}, \mathcal{F}, \text{LUT}^{\nn}_{\text{Power}}, P_{\text{max}}}$ \Comment{Based on \Cref{eq:feasible_combination}}
    \State $TT^{p}_{\text{acc}} \gets TT_{\text{acc}} [C]$
    \State $(b\_\text{idx}, f\_\text{idx})\gets \textbf{argmin} (TT^{p}_{\text{acc}})$
    \State $b^{*}, f^{*} \gets \mathcal{B}[b\_\text{idx}], \mathcal{F}[f\_\text{idx}]$ \Comment{Selected batch size and GPU frequency}

\end{algorithmic}

%% file: results.tex
\section{Results}
\label{sec:results}
In this section, we evaluate the training time and energy consumption for finetuning vision and text tasks on Nvidia Jetson Nano and TX2NX. Additionally, we assess the effectiveness of our approach by conducting a sensitivity analysis on the selection of the proxy dataset.

\subsection{Experiment Setup}

\paragraph{Datasets and models:}
We evaluate our approach for the image object classification and next-word prediction tasks. For image classification, a model is pre-trained with the full CIFAR100 \cite{cifars_cite} and to be trained on subsets (i.e., quarter) of SVHN \cite{SVHN_cite} and CINIC \cite{cinic_cite} datasets on the device. We use a subset of CIFAR10 \cite{cifars_cite} as the proxy dataset on the server. For all datasets
, the input images are of size 3 $\times$ 32 $\times$ 32. Each dataset (subset) is divided into training and testing sets with an 80/20 split, with target accuracy evaluated on the test split. We evaluate ResNet18 \cite{resnet_cite} and MobileNetV2 \cite{mobilenet_cite} models (widely adopted on edge devices), where both of them are trained with Adam optimizer \cite{adam_cite}. A summary of the experimental settings is provided in \Cref{table:experimet_setting}.

For the next character prediction task, we evaluate a 6 layer transformers with 6 attention heads per attention block, 256 embedding dimensions, and a sequence length of 64. We use AdamW \cite{Adamw_cite} as an optimizer. We pre-train the model on WikiText-2 dataset \cite{wiki_text_cite}, utilize tiny shakespeare \cite{tiny_shakespeare} as a proxy dataset, and train it on some Jane Austin and Charles dickens novels\footnote{We used the text for works of Jane Austin's from nltk package \cite{nltk_library_cite} and downloaded Dickens works from project Gutenburg. More details are provided in \cref{appendix:text_datasets}}. We use $90\%$ of a dataset for training and the rest for testing.  For all datasets, we fix the vocabulary to include only words with English letters, digits, punctuation, spaces, and new lines. We set the target character level accuracy for Austin and Dickens at 62\% and 61\%, respectively. 

\input{experiments_setting_table}

\input{power_limit_cifar_proxy_std_2}

\paragraph{Batch size and learning rate:} The choice of appropriate learning rate and batch size are often intertwined, as they impact each other’s effectiveness. Larger batch sizes provide more stable gradient estimates, potentially permitting the use of higher learning rates. 
Therefore, to preserve the performance of deep models with different batch sizes, we apply learning rate scaling (i.e., square root scaling \cite{krizhevsky2014one} for Adam and AdamW). For ResNet18, the batch sizes ranged from~4 to~128, consisting exclusively of powers of two. The initial learning rate of~$5 \times 10^{-4}$ is used for the largest batch size of~128 (with learning rates scaled for other batch sizes). The same setup was also applied to MobileNetV2; however, the batch size of~128 was omitted due to memory constraints.
For transformers, we similarly consider batch sizes of 4 to 128 with a learning rate of $1 \times 10^{-3}$ for the batch size of 128.

\paragraph{Hardware and power limits:} 

We evaluate our method on Nvidia Jetson Nano with \qty{4}{\giga\byte} memory on three scenarios. In the first and second scenarios, the power limits at the device are set to $P_{\text{max}}^{1} = \SI{4.5}{\watt}$ and $P_{\text{max}}^{2} = \SI{7}{\watt}$. In the third, the device operates without any power limits, denoted as N/A.  To show that our solution and our results are not device specific, we also provide evaluation on another device (i.e., Nvidia Jetson TX2NX). We use PyTorch 1.10 \cite{pytorch_cite} for Jetson Nano and TX2NX. For pretraining and proxy datasets' training we use NVIDIA A6000 GPU with Pytorch 2.1.

\paragraph{Comparison baselines:} We compare our approach to the following baselines that depend on state of the art techniques:
\begin{itemize}

    \item \textbf{Baseline 1:} The state of practice is to use the largest~$b$  that can fit into memory \cite{goyal2017accurate, ecsa-10-16202}, where the latter use edge GPUs. For the three power limits, we use $\SI{307}{\mega\hertz}$,  $\SI{614}{\mega\hertz}$, and $\SI{921}{\mega\hertz}$ as upper-bound operating GPU frequencies for the device. These are determined based on profiling training of different models and selecting the frequencies that assure a power limit is satisfied irrespective of what model or $b$ is used for the training.

    \item \textbf{Baseline 2:} We select the value of~$b$ that minimizes~$N_{{s}_{\text{acc}}}$ on the proxy dataset, but we use the same GPU frequencies as in Baseline 1, so no joint optimization is applied. 
    \item \textbf{Fastest configuration:} This baseline serves as an \textit{upper bound} where optimal~$f$ and~$b$ are selected. To determine this, we train the given model on the target dataset on device with all batch sizes
to get~$N_{s_{\text{acc}}}$, substitute in \cref{eq:time_to_acc2} and finally select the best $b$ and $f$ configuration on the device. 
\end{itemize}
We repeat every experiment with different five seeds for image classification tasks and three seeds for next character prediction and record the mean and standard deviation.

\input{power_limit_transformers_pretrained}

\subsection{Training Time Evaluation}
\label{subsec:training_time_evaluation}

\Cref{table:comparison_with_baseline_nano} and \Cref{table:comparison_with_baseline_nano_transformers} report the training time comparison of our approach and the three baselines on Nvidia Jetson Nano. \Cref{table:comparison_with_baseline_tx2nx} provides additional comparison for image classification task on Nvidia Jetson TX2NX.

Firstly, we start with the evaluation of image classification tasks reported in \Cref{table:comparison_with_baseline_nano}. For ResNet18 training, our method consistently outperforms baseline~1, achieving~$1.5-2.4\times$ reduction in training time. It also outperforms baseline~2 with~$1.4-1.7\times$ less training time. These two baselines are only able to explore a small subset of the design space, potentially missing the optimal configuration. 

Finally, compared to the fastest configuration, our method selects the same configurations in most of the evaluations, while in the case of a different selection, a minor performance decline (of $3\%$) is observed. 

Training of MobileNetV2 has three distinct characteristics: lower peak power, better parallelization for larger batch sizes, and different ~$\mathbf{r}$. 
Our method adapts to that by selecting larger batch sizes (and frequencies) than those selected for ResNet18. Compared with baseline~1, our method performs better with minor differences in training time since both select large batch sizes and maximum $f$ utilizing the power.
In low-power, the gain from our approach increases as the selected batch size enables higher frequencies, reducing training time by up to~$1.4\times$. Baseline~2 misses the opportunity to use higher frequencies, especially as MobileNetV2 consumes less power. Ultimately, our method selects the fastest configurations when training MobileNetV2 in all power-limits scenarios.

Next, we examine the next character prediction training. As shown in \Cref{table:comparison_with_baseline_nano_transformers}, our method improves the training time by $1.15 \times$ and $2.14\times$ compared to baseline $1$ and $2$, while choosing the same configurations as the fastest configuration.

In \Cref{table:comparison_with_baseline_tx2nx}, we compare our solution with the baselines for ResNet18 training on Nvidia Jetson TX2NX, where the device receives $\mathbf{r}$ for ResNet18 training from the server. For training on SVHN, our method outperforms both the baselines $1$ and $2$, reducing the training time by $1.7\times$ and $1.8\times$, respectively. Furthermore, it selects the same configuration as the optimal one. Similarly, when training on the CINIC dataset, baselines $1$ and $2$ require up to $1.94\times$ and $2.04\times$ more training time compared to ours. Our method did not select the exact optimal configuration in this scenario; however, it selects a near-optimal configuration that results in only $2\%-8.2\%$ more training time compared to optimal.  

In summary, our method outperforms existing baselines, reducing training time to accuracy across various model architectures, tasks, and hardware.

\input{power_limit_tx2_resnet18}

\begin{figure*}[t!]
    \centering
    \begin{subfigure}[b]{\textwidth}
    \hspace{16mm}
    \includegraphics[]{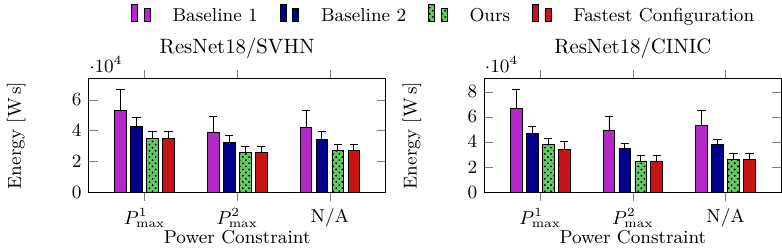}
    \end{subfigure}
    \vskip \baselineskip \hspace{1mm} 
    \begin{subfigure}[b]{0.40\textwidth}
        \includegraphics[]{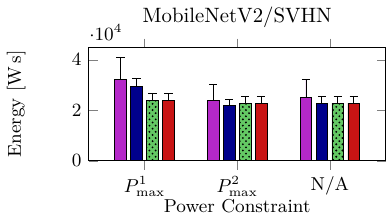}
    \end{subfigure}
    \begin{subfigure}[b]{0.40\textwidth}
        \includegraphics[]{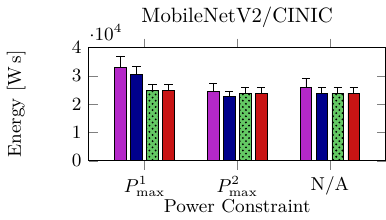}
    \end{subfigure}
    
    \vskip \baselineskip \hspace{1mm} 
    \begin{subfigure}[b]{0.40\textwidth}
        \includegraphics[]{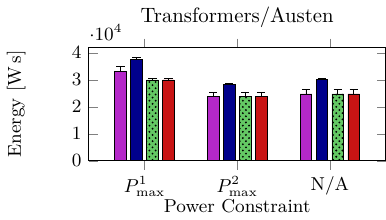}
    \end{subfigure}
    \begin{subfigure}[b]{0.40\textwidth}
        \includegraphics[]{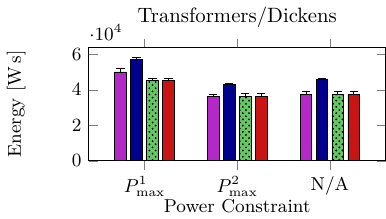}
    \end{subfigure}

    \caption{Energy consumption comparison between our approach and baseline methods during training under three power constraint scenarios. The recorded data includes training ResNet18 and MobileNetV2 on the SVHN and CINIC datasets, as well as a transformers network on the Austen and Dickens dataset, all performed on an Nvidia Jetson Nano.}
    \label{fig:energy_comparison}
\end{figure*}

\subsection{Energy Evaluation}
We investigate the energy efficiency of our proposed approach for the image object classification and next character prediction tasks in \Cref{fig:energy_comparison}. 

For the image classification task, it is noticeable that our method always outperforms baseline~1 since it takes less training time given the power limits. In addition, selections made by baselines~1 and~2 can lead to up to~$2\times$ and~$1.4\times$ more energy usage compared to our method, respectively. It is important to state that minimizing the training time does not always mean minimizing energy consumption. An example is comparing the energy consumption for~$P_{\text{max}}^{2}$ on ResNet18 with the no power limit N/A. Despite having lower training time in N/A (as shown in \Cref{table:comparison_with_baseline_nano}), training under~$P_{\text{max}}^{2}$ is more energy efficient since the used frequencies for~$P_{\text{max}}^{2}$ provide a better tradeoff between performance and power. The same applies for Baseline~2 training MobileNetV2 under ~$P_{\text{max}}^{2}$, that uses a lower frequency with the same batch size of our approach. 

For the next character prediction task, our approach uses $9\%$ less energy for $P_\text{{max}}^{1}$ and nearly the same energy consumption for $P_\text{{max}}^{2}$ while training in less time (the same configuration selection for $P_\text{{max}}^{3}$) compared to the baseline $1$. In addition, our method records $1.25\times$ less training energy for the baseline $2$.

\begin{figure*}
    \centering
    \resizebox{\textwidth}{!}{
    \begin{subfigure}[b]{0.6\textwidth}
        \includegraphics[]{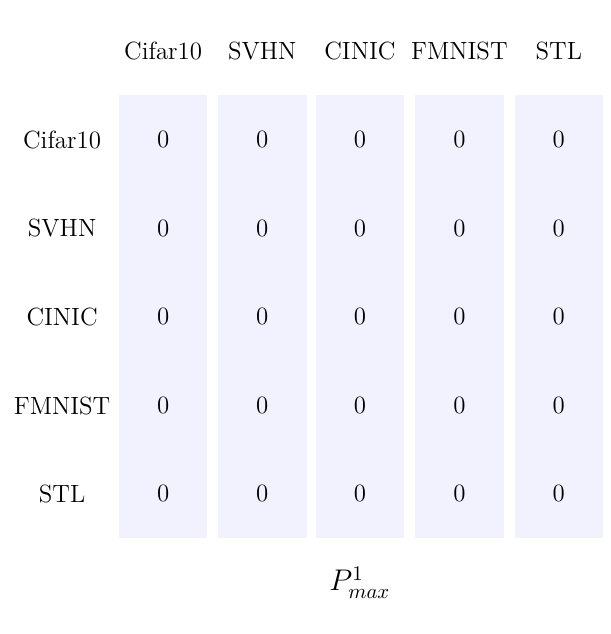}
        \end{subfigure}
        \hspace{0.3mm}
        \begin{subfigure}[b]{0.6\textwidth}
            \includegraphics[]{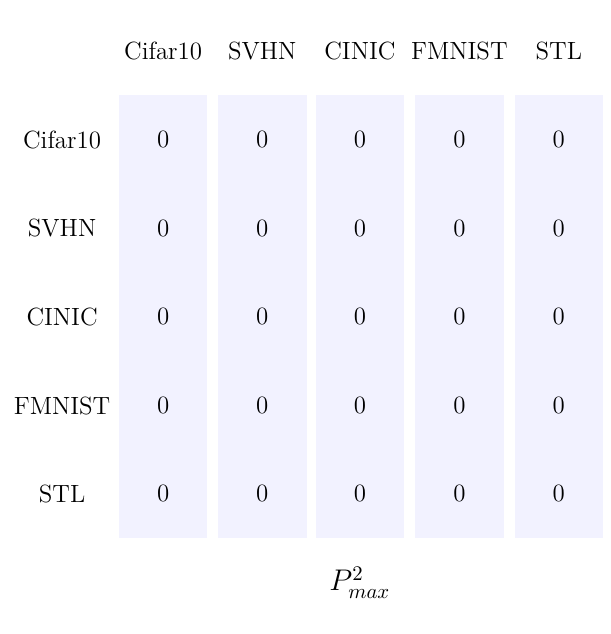}
        \end{subfigure}
        \hspace{0.3mm}
        \begin{subfigure}[b]{0.6\textwidth}
            \includegraphics[]{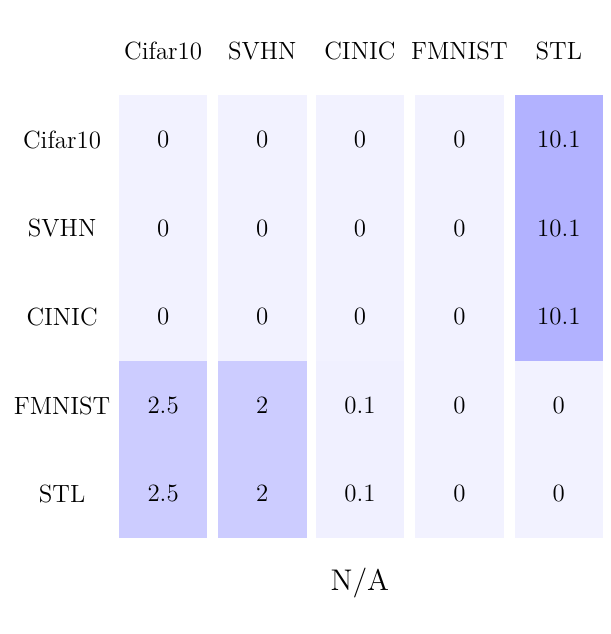}
        \end{subfigure}
    }
    \caption{Confusion Matrix of time increase percentages to the fastest configuration for image classification datasets on Nvidia Jetson Nano across three power constraints for MobileNetV2 training. The rows represent the selected proxy dataset while the columns represent the target datasets where training on edge is conducted on. The results indicate that the proposed method is not sensitive to the selection of proxy dataset. }
    \label{fig:confusion_matrix_mobilenet}
\end{figure*}
\begin{figure*}
    \centering
    \resizebox{0.85\textwidth}{!}{
        \begin{subfigure}{0.29\textwidth}
        \includegraphics[]{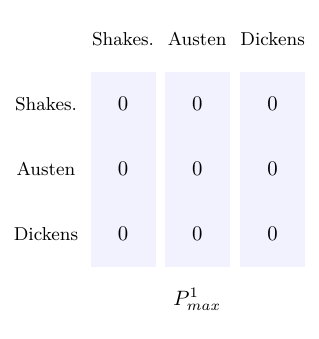}
        \end{subfigure}
        \hspace{2mm}
        \begin{subfigure}{0.29\textwidth}
            \includegraphics[]{transformer_p1.pdf}
        \end{subfigure}
        \hspace{2mm}
        \begin{subfigure}{0.29\textwidth}
           \includegraphics[]{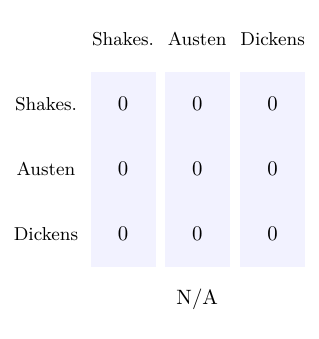}
        \end{subfigure}
    }
    \caption{ Confusion Matrix of time increase percentages to the fastest configuration for next character prediction on Nvidia Jetson Nano across three power constraints for transformers training. The rows represent the selected proxy dataset while the columns represent the target datasets where training on edge is conducted on. The results indicate that the proposed method is not sensitive to the selection of proxy dataset.}
    \label{fig:confusion_matrix_transformers1}
\end{figure*}

\subsection{Proxy Dataset Analysis}

We conduct a sensitivity analysis for the selection of the proxy dataset for image classification and next character prediction tasks by testing all possible proxy and target dataset combinations. 

For the image classification task, we add two additional datasets namely, Fashion MNIST \cite{fashion_mnist_cite} (FMNIST)  and STL \cite{STL_cite}. The tasks include object, digit, and fashion classification, with dataset sizes ranging from $5K$ to $22.5K$ samples. We provide in \cref{fig:confusion_matrix_mobilenet} a confusion matrix showing the percentage of time increase for the choice of proxy dataset selection compared to the fastest configuration for different datasets for training MobileNetV2 on Nvidia Jetson Nano. The results show that our approach selects the correct configurations in most of the cases with only a few instances where the optimal configuration is not selected in no power constraint scenario. The same observation is also applied to ResNet18 as shown in \Cref{fig:confusion_matrix_resnet18} (\cref{appendix:proxy_dataset_extension}). For the next character prediction, we provide the confusion matrix in \Cref{fig:confusion_matrix_transformers1}. The optimal configuration is selected regardless of the select proxy dataset.

To sum up, the results show that our approach is not sensitive to the selected proxy dataset, demonstrating the practicality of the proposed solution.

%% file: experiments_setting_table.tex
\begin{table}[tb!]
  \caption{Experimental setup for image classification datasets and models used.}
  \label{table:experimet_setting}
  \centering
  \begin{tabular}{lccc}
    \toprule
    Model        & Dataset & Dataset size & Target Acc. \\
    \midrule
    ResNet18     & Cifar10 & 12500        &  87\% \\
    ResNet18     & SVHN    & 18315        &  92\% \\
    ResNet18     & CINIC   & 22500        &  76\% \\
    \midrule
    MobileNetV2  & Cifar10 & 12500        &  84.5\% \\
    MobileNetV2  & SVHN    & 18315        &  91.5\% \\
    MobileNetV2  & CINIC   & 22500        &  74\% \\
    \bottomrule
  \end{tabular}
\end{table}

%% file: power_limit_cifar_proxy_std_2.tex
\begin{table*}[thb!]
  \caption{Training time comparison with baselines and upper bound over three different power limits (i.e., $P_\text{{max}}^{1}$, $P_{\text{max}}^{2}$, and N/A). Recorded times are in seconds. CIFAR10 is used as proxy dataset in our proposed approach. (Evaluation time is excluded)}
  \label{table:comparison_with_baseline_nano}
  \centering
  \begin{adjustbox}{width=\columnwidth,center}
  \begin{tabular}{lllllllllll}
    \toprule
    Model       & Method                & \multicolumn{3}{c}{SVHN}          & \multicolumn{3}{c}{CINIC} \\
    \cmidrule(r){3-5} \cmidrule(r){6-8}
                &                       & $P_\text{{max}}^{1}$ &$P_\text{{max}}^{2}$&N/A  &$P_{\text{max}}^{1}$&$P_{\text{max}}^{2}$& N/A \\
    \midrule
    ResNet18    & Baseline 1            &14347$\pm$3796  &8204$\pm$2170    &6187$\pm$1637   &18154$\pm$4109   &10381$\pm$2349 &7829$\pm$1772 \\
    ResNet18    & Baseline 2            &11345$\pm$1616 &7055$\pm$1005    &5817$\pm$\phantom{0}828   &12436$\pm$1443    &\phantom{0}7733$\pm$\phantom{0}897   &6377$\pm$\phantom{0}740 \\
    ResNet18    & Ours                  &\phantom{0}\textbf{8477$\pm$1207}  &\textbf{4607$\pm$\phantom{0}725}    &\textbf{4170$\pm$\phantom{0}656}    &\phantom{0}\textbf{9292$\pm$1078}    &\textbf{\phantom{0}4454$\pm$\phantom{0}788}   &\textbf{4032$\pm$713}  \\
    ResNet18    & Fastest configuration & \phantom{0}8477$\pm$1207  &4607$\pm$\phantom{0}725    &4170$\pm$\phantom{0}656  &\phantom{0}9030$\pm$1598   &\phantom{0}4454$\pm$\phantom{0}788   &4032$\pm$\phantom{0}713  \\
    \midrule
    
    MobileNetV2 & Baseline 1            &\phantom{0}8708$\pm$2396  &5107$\pm$1405     &3990$\pm$1098   &\phantom{0}8909$\pm$1121    &\phantom{0}5226$\pm$\phantom{0}657    &4082$\pm$\phantom{0}513  \\
    MobileNetV2 & Baseline 2            &\phantom{0}8082$\pm$\phantom{0}912  &4871$\pm$\phantom{0}550     &3912$\pm$\phantom{0}441   &\phantom{0}8428$\pm$\phantom{0}692     &\phantom{0}5079$\pm$\phantom{0}417    &4080$\pm$\phantom{0}335 \\
    MobileNetV2 & Ours                  &\phantom{0}\textbf{5940$\pm$\phantom{0}916}   &\textbf{3912$\pm$\phantom{0}441}     &\textbf{3912$\pm$\phantom{0}441}   &\phantom{0}\textbf{6194$\pm$\phantom{0}509}     &\textbf{\phantom{0}4080$\pm$\phantom{0}335}    &\textbf{4080$\pm$\phantom{0}335}  \\
    MobileNetV2 & Fastest configuration &\phantom{0}5940$\pm$\phantom{0}916   &3912$\pm$\phantom{0}441    &3912$\pm$\phantom{0}441   &\phantom{0}6194$\pm$\phantom{0}509     &\phantom{0}4080$\pm$\phantom{0}335    &4080$\pm$\phantom{0}335  \\
    \bottomrule
  \end{tabular}
  \end{adjustbox}
\end{table*}

%% file: power_limit_transformers_pretrained.tex
\begin{table*}[tb!]
  \caption{Training time comparison with baselines and upper bound for different power limits (i.e., $P_\text{{max}}^{1}$, $P_{\text{max}}^{2}$, and N/A). Recorded times are in seconds. Tiny Shakespeare is used as proxy dataset in our approach. (Evaluation time is excluded)}
  \label{table:comparison_with_baseline_nano_transformers}
  \centering
  \begin{adjustbox}{width=\columnwidth,center}
  \begin{tabular}{lllllllllll}
    \toprule
    Model       & Method                & \multicolumn{3}{c}{Austen}          & \multicolumn{3}{c}{Dickens} \\
    \cmidrule(r){3-5} \cmidrule(r){6-8}
                &                       & $P_\text{{max}}^{1}$ &$P_\text{{max}}^{2}$&N/A  &$P_{\text{max}}^{1}$&$P_{\text{max}}^{2}$& N/A \\
    \midrule

    Transformer    & Baseline 1            &8136$\pm$524   &4607$\pm$296   &3556$\pm$229     &12347$\pm$439   &6991$\pm$249   &5397$\pm$192 \\
    Transformer    & Baseline 2            &9425$\pm$131   &6081$\pm$85    &5193$\pm$72      &14278$\pm$228  &12941$\pm$532  &7867$\pm$125 \\
    Transformer    & Ours                  &\textbf{7232$\pm$101}   &\textbf{3982$\pm$256}   &\textbf{3556$\pm$229}     &\textbf{10956$\pm$175}   &\textbf{6043$\pm$215}   &\textbf{5397$\pm$192}  \\
    Transformer    & Fastest configuration &7232$\pm$101   &3982$\pm$256   &3556$\pm$229     &10956$\pm$175   &6043$\pm$215   & 5397$\pm$192  \\
    \bottomrule
  \end{tabular}
  \end{adjustbox}
\end{table*}

%% file: power_limit_tx2_resnet18.tex
\begin{table*}[tb!]
  \caption{Training time comparison with baselines and upper bound over three different power limits (i.e., $P_\text{{max}}^{1}$, $P_{\text{max}}^{2}$, and N/A) on Nvidia Jetson TX2NX. Recorded times are in seconds. CIFAR10 is used as proxy dataset in our proposed approach. (Evaluation time is excluded)}
  \label{table:comparison_with_baseline_tx2nx}
  \centering
  \begin{adjustbox}{width=\columnwidth,center}
  \begin{tabular}{lllllllllll}
    \toprule
    Model       & Method                & \multicolumn{3}{c}{SVHN}          & \multicolumn{3}{c}{CINIC} \\
    \cmidrule(r){3-5} \cmidrule(r){6-8}
                &                       & $P_\text{{max}}^{1}$ &$P_\text{{max}}^{2}$&N/A  &$P_{\text{max}}^{1}$&$P_{\text{max}}^{2}$& N/A \\
    \midrule
    ResNet18    & Baseline 1            &7343$\pm$1943  &3944$\pm$1043    &3021$\pm$799   &9292$\pm$2103   &4991$\pm$1129 &3823$\pm$865 \\
    ResNet18    & Baseline 2            &7337$\pm$1943 &4771$\pm$679    &4086$\pm$582   &8042$\pm$933    &5230$\pm$\phantom{0}897   &4479$\pm$520 \\
    ResNet18    & Ours                  &\textbf{4283$\pm$708}  &\textbf{2648$\pm$417}    &\textbf{2431$\pm$382}    &\textbf{5463$\pm$889}    &\textbf{2560$\pm$453}   &\textbf{2350$\pm$416}  \\
    ResNet18    & Fastest configuration &4283$\pm$708  &2648$\pm$417    &2431$\pm$382  &5046$\pm$893   &2510$\pm$\phantom{0}625   &2294$\pm$571  \\
    \bottomrule
  \end{tabular}
  \end{adjustbox}
\end{table*}

%% file: conclusion.tex
\section{Conclusion}
In this work, we propose a power-aware training method aimed at accelerating training on power-constrained GPU devices. Our results show that great savings can be achieved in terms of training time and energy consumption, when carefully and jointly selecting the system and application parameters for training. This is achieved without scarifying the training model quality. 
The proposed solution is applicable to a wide range of models (including, but not limited to, CNNs and transformers).

Sustainability is one of the major issues when it comes to large scale adoption of AI services. Our solution can be employed to make training over such systems more energy efficient, reducing its carbon footprint. It can also be used to integrate the renewable energy resources deployed near the edge devices, making the training process even greener, i.e., by adapting the GPU frequency and batch size at the devices to the level of available green energy. 

A limitation of this work is that we assume that the power constraint on a device is constant and does not change through the training period. For longer training jobs that can take multiple days, this assumption will not hold. For future work, we want to extend our study to evaluate varying power constraints and distributed learning settings where heterogeneous devices with different constraints contribute to the training.

%% file: appendix.tex
\clearpage
\appendix

\section{Appendix}

\subsection{Training on edge CPU}
\label{appendix:CPU_training}
\begin{figure}[htb!]
    \centering
    \includegraphics[]{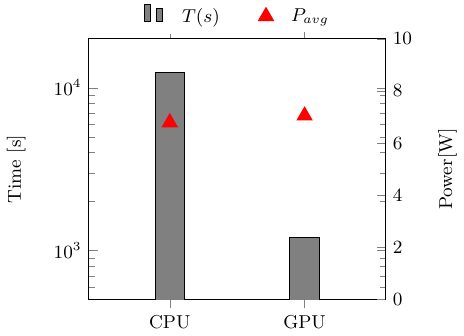}
    \label{fig:cpu_gpu_latency}
    \caption{ResNet18 time for training 50K samples ($T_s$) and average power (red triangles) of Cifar10 on Jetson Nano over CPU and GPU.}
 
\end{figure}

In this section, we briefly discuss training on an edge device's CPUs (i.e., Jetson Nano) and the reasons of why to focus only on training on GPUs on such devices. In \cref{fig:cpu_gpu_latency}, we show the processing time and power for training over 50K samples with batch size 32, where both CPU and GPU are using the maximum frequency. We observe that the GPU is an order of magnitude faster compared to the CPU, while nearly consuming similar average power. Similarly, the GPU would be nearly an order of magnitude more energy efficient on these devices compared to the CPU. Thus, we focus in our work on edge GPUs as training is more time and energy efficient.  

\subsection{Proxy Dataset Analysis}
\label{appendix:proxy_dataset_extension}

\begin{figure*}
    \centering
    \resizebox{\textwidth}{!}{
    \begin{subfigure}[b]{\textwidth}
        \includegraphics[]{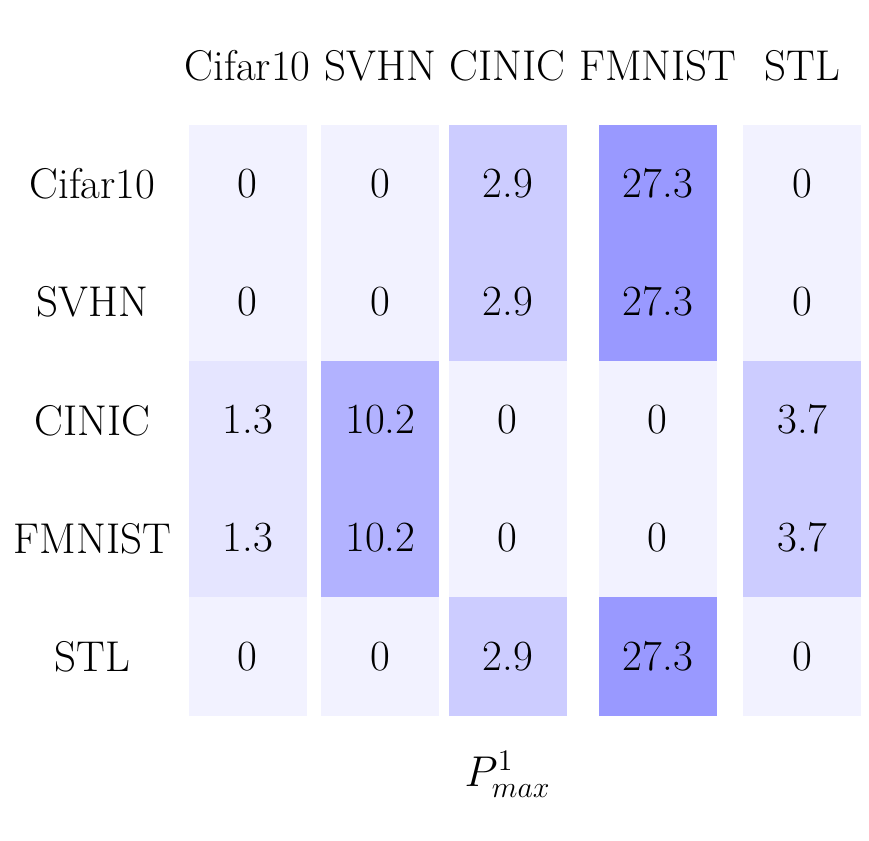}
    \end{subfigure}
    \hspace{1mm}
    \begin{subfigure}[b]{\textwidth}
        \includegraphics[]{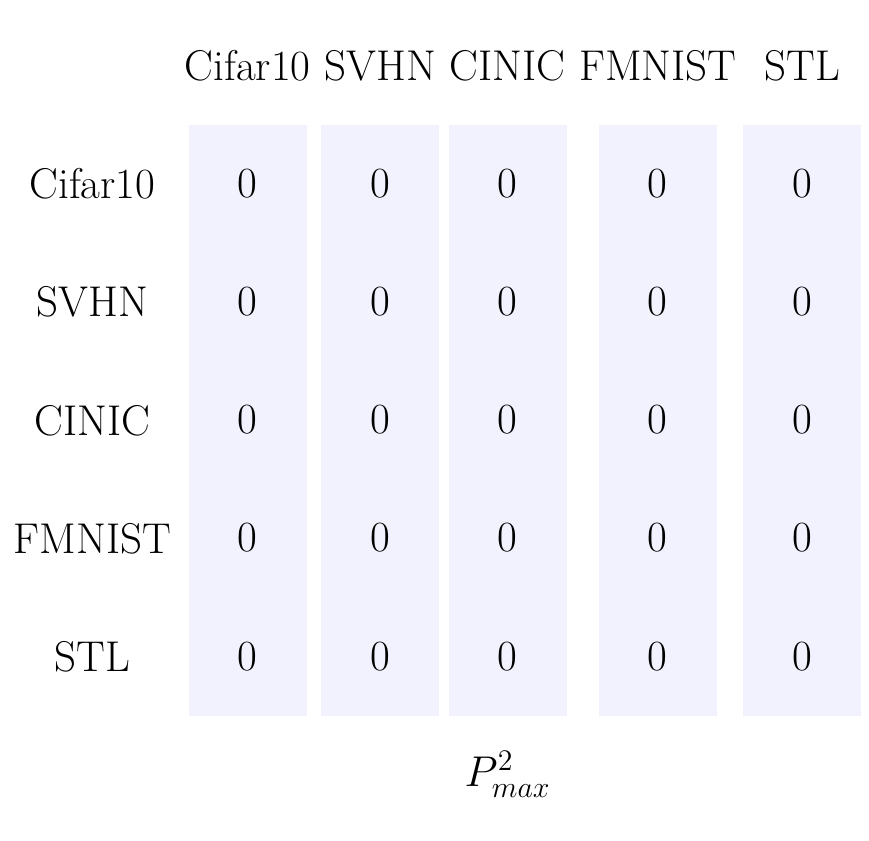}
    \end{subfigure}
    \hspace{1mm}
    \begin{subfigure}[b]{\textwidth}
        \includegraphics[]{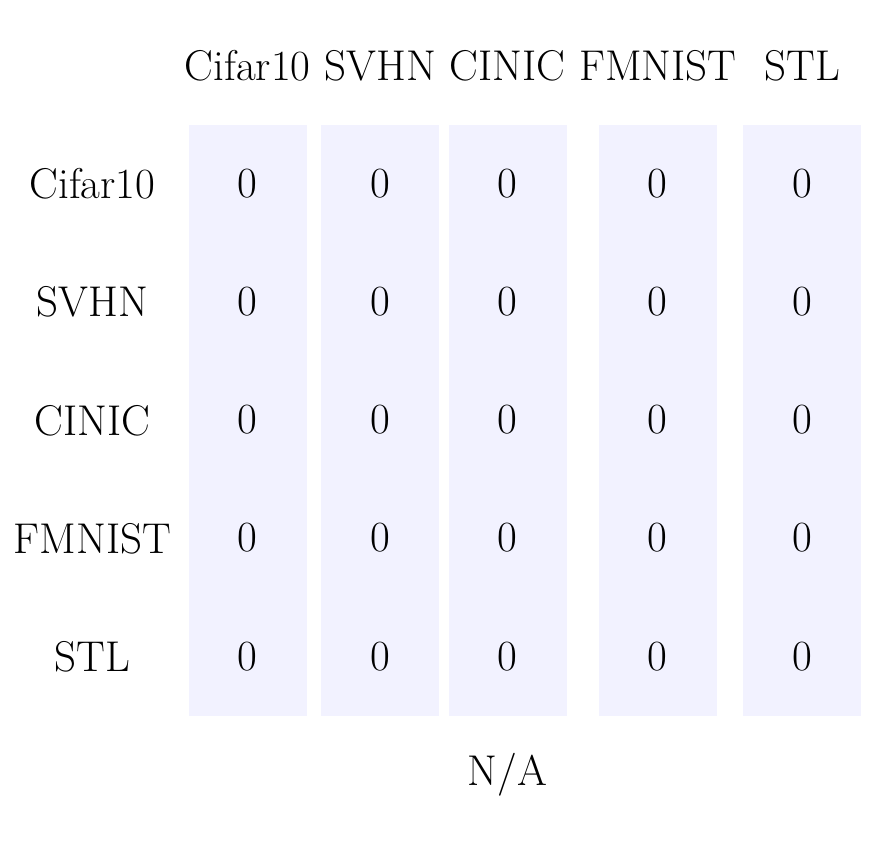}
    \end{subfigure}
    }
    \caption{Confusion Matrix of time increase percentages to the fastest configuration for image classification datasets across three power constraints for ResNet18 training.}
    \label{fig:confusion_matrix_resnet18}
\end{figure*}

We provide a continuation for the sensitivity analysis for the proxy dataset selection. In \cref{fig:confusion_matrix_resnet18}, provide the confusion matrix for selecting different proxy datasets to train ResNet18. The difference for ResNet18 compared to MobileNetV2 (in \Cref{fig:confusion_matrix_mobilenet})is that the suboptimal configurations are selected for the lowest power constraint, while for other cases the selected configurations match the optimal.

\subsection{Text datasets details}
\label{appendix:text_datasets}
For Jane Austen's dataset part, we extracted it from the nltk package (gutenberg corpus) where the text consists of 3 novel namely Emma, Persuasion, and Sense and Sensibility.  For Dickens, we used eight novels namely The Pickwick Papers, Pictures from Italy,  A Tale of Two Cities, A Story of the French Revolution, The Chimes, Mugby Junction, The Haunted Man and the Ghost's Bargain, and The Mystery of Edwin Drood.
We filter the licensing text, author history, etc. from the training and testing text.
    
\subsection{Power measurement details}

We read the power sensor values during training with a sample rate of 1\si{\second}. For the Nvidia Jetson Nano, the power measurements are available to read from using sysfs nodes. For example, on the jetson nano, the sysfs nodes are available in the path '/sys/bus/i2c/drivers/ina3221x/6-0040/iio:device0/'.